\documentclass[lettersize,journal]{IEEEtran}
\usepackage{subfig}
\usepackage{picins}
\usepackage{times}
\usepackage{soul}
\usepackage{url}
\usepackage[hidelinks]{hyperref}
\usepackage[utf8]{inputenc}
\usepackage{caption}
\usepackage{graphicx}
\usepackage{amsmath}
\usepackage{amsthm}
\usepackage{booktabs}
\usepackage{algorithm}
\usepackage{algorithmic}
\usepackage[switch]{lineno}
\usepackage{multirow}
\usepackage{textcomp}
\usepackage{stfloats}
\usepackage{verbatim}
\usepackage{amsmath,amsfonts}
\usepackage{cite}
\begin{document}

\title{PointCore: Efficient Unsupervised Point Cloud Anomaly Detector Using  Local-Global Features}

\author{Baozhu Zhao, Qiwei Xiong, Xiaohan Zhang, Jingfeng Guo, Qi Liu,~\IEEEmembership{Senior Member,~IEEE,} Xiaofen Xing,~\IEEEmembership{Member,~IEEE,} Xiangmin Xu,~\IEEEmembership{Senior Member,~IEEE}
\thanks{Corresponding author: Qi Liu (drliuqi@scut.edu.cn)}
\thanks{B. Zhao, Q. Xiong, X. Zhang, J. Guo and X. Xu are with the School of Future Technology, South China University of Technology, China 511400 }
\thanks{X. Xing is with the School of Electronic and
Information Engineering, South China University of Technology, Guangzhou 510640, China.}
}

\maketitle

\begin{abstract}
    Three-dimensional point cloud anomaly detection that aims to detect anomaly data points from a training set serves as the foundation for a variety of applications, including industrial inspection and autonomous driving. However, existing point cloud anomaly detection methods often incorporate multiple feature memory banks to fully preserve local and global representations, which comes at the high cost of computational complexity and mismatches between features. To address that, we propose an unsupervised point cloud anomaly detection framework based on joint local-global features, termed PointCore. To be specific, PointCore only requires a single memory bank to store local (coordinate) and global (PointMAE) representations and different priorities are assigned to these local-global features, thereby reducing the computational cost and mismatching disturbance in inference. Furthermore, to robust against the outliers, a normalization ranking method is introduced to not only adjust values of different scales to a notionally common scale, but also transform densely-distributed data into a uniform distribution. Extensive experiments on Real3D-AD dataset demonstrate that PointCore achieves competitive inference time and the best performance in both detection and localization as compared to the state-of-the-art Reg3D-AD approach and several competitors.
    
\end{abstract}

\begin{IEEEkeywords}
3D point cloud, anomaly detection, unsupervised learning
\end{IEEEkeywords}

\section{Introduction}
\IEEEPARstart{A}{nomaly} detection aims to find the abnormal region of products and plays an important role in various fields, such as industrial quality inspection \cite{Carrera_Manganini_Boracchi_Lanzarone_2017,Song_Yan_2013}, and autonomous driving \cite{Hendrycks_Basart_Mazeika_Mostajabi_Steinhardt_Song_2019,xu2020predictive}. Current anomaly detection methods \cite{Zavrtanik_Kristan_Skočaj_2021,zhou2019attention,10415065} are mostly unsupervised and they target on two-dimensional (2D) images, where models are typically trained on images with well-studied architectures. For 3D point cloud-based anomaly detection task, there remains relatively unexplored within the current literature. Compared with 2D images, 3D point clouds have the advantage of richer structural information, which, however, gets along with but disorder, highly sparse and irregular distribution. To process point clouds distinctly, various handcrafted or deep learning-based feature descriptors at different scales are applied.
\begin{figure}
    \centering
    \includegraphics[width=0.9\linewidth]{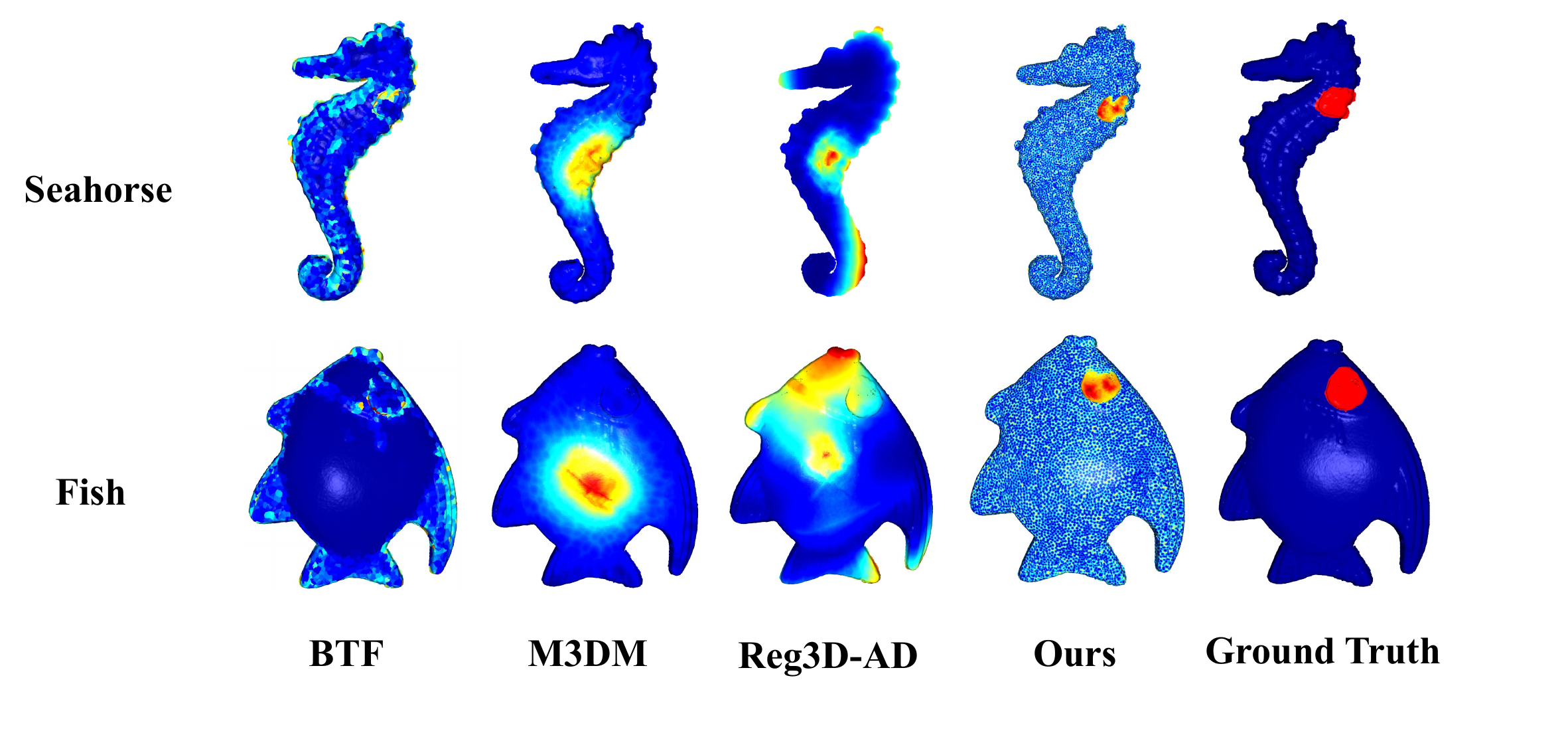}
    \caption{Heatmaps of anomaly scores obtained by several methods on the Real3D-AD dataset. From the visualization, we see that the proposed method can detect and locate anomaly data points more accurately compared to others.}
    \label{fig:enter-label}
\end{figure}

Recently, a large-scale, high-resolution 3D anomaly detection dataset, termed Real3D-AD \cite{Liu_Xie_Chen_Li_Wang_Liu_Wang_Zheng}, was provided. The objects in Real3D-AD dataset have 0.001mm-0.0015mm resolution, 360 degree coverage and perfect prototype. In \cite{Liu_Xie_Chen_Li_Wang_Liu_Wang_Zheng}, the authors applied PatchCore \cite{Roth_Pemula_Zepeda_Scholkopf_Brox_Gehler_2022} from image anomaly detection to point cloud anomaly detection, and developed a general-purposed registration-based point cloud anomaly detector, dubbed as Reg3D-AD. Reg3D-AD employs a dual-feature representation approach to preserve the local and global features of training prototypes, which achieves remarkable detection accuracy but slow inference. Current point cloud anomaly detectors can be categorized into twofolds: (1) Reconstruction-based methods, which reconstruct the input point cloud data via autoencoder \cite{Li_Xu_2023,zhu2020view,peng2023class} and identify anomalies by comparing the deviation between the original and reconstructed data. However, these approaches are sensitive to the resolution of point clouds, leading to slower inference speed and worse accuracy. (2) Memory bank-based approaches \cite{Liu_Xie_Chen_Li_Wang_Liu_Wang_Zheng,xiao2023distinguishing}, where memory banks are useful for storing representative features to implicitly build a normal distribution and looking for the out-of-distributed defects. Compared with the former, using a pretrained feature extractor directly to construct a memory bank enjoys fast training speed and is not affected by the resolution of point clouds. Besides that, existing point cloud anomaly detectors \cite{Liu_Xie_Chen_Li_Wang_Liu_Wang_Zheng,Horwitz_Hoshen_2022} often incorporate multiple feature memory banks to fully preserve local and global representations, which comes at the high cost of computational complexity and mismatching between features.

To address the above issues, we propose an unsupervised point cloud anomaly detection framework based on joint local-global features, termed PointCore. To be specific, our contributions are summarized as follows:
\begin{itemize}
    \item [1.]
    PointCore only requires a  memory bank to store local-global representations and different priorities are assigned to these local-global features to reduce the computational cost and mismatching disturbance for inference. 
    \item[2.]
    We propose a ranking-based normalization method to eliminate the distribution differences among various anomaly scores and apply the point-to-plane Iterative Closest Points(point-plane ICP) algorithm to perform local optimization of point cloud registration results for robust decision-making.
    \item[3.]
    Extensive experiments on Real3D-AD dataset demonstrate that PointCore achieves competitive inference time and the best performance in both detection and localization as compared to the state-of-the-art Reg3D-AD approach and several competitors.
\end{itemize}

\renewcommand{\dblfloatpagefraction}{.8}
\begin{figure*}
    \centering
    \includegraphics[width=0.8\linewidth]{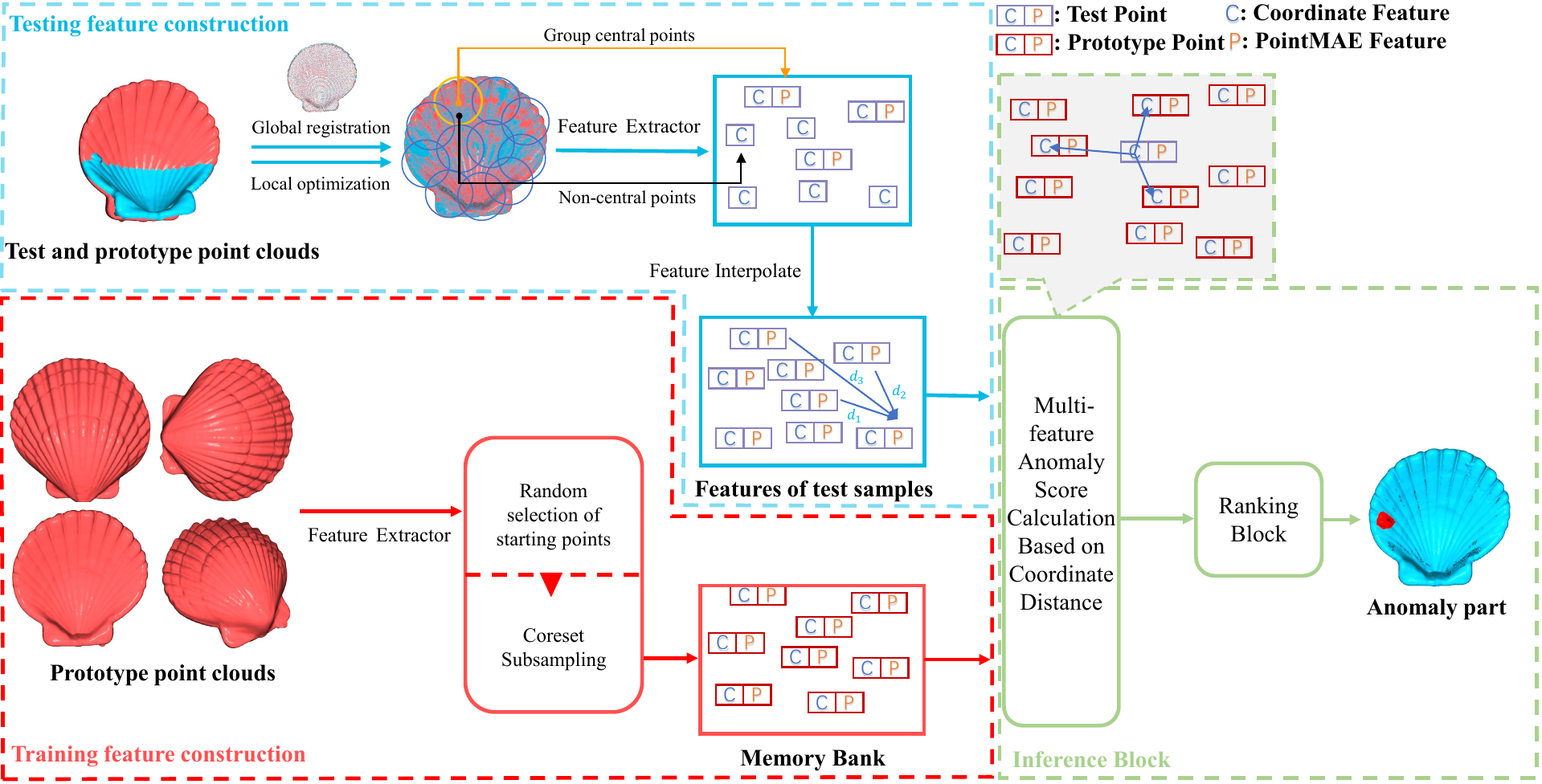}
    \caption{\textbf{The pipeline of PointCore architecture}. We randomly select and convert template point clouds as the reference coordinates through global registration and local optimization methods, use  pre-trained PointMAE  feature extractor to obtain the PointMAE features, and then bind the coordinates and the features to establish the coordinate-PointMAE memory bank. Finally, we compute anomaly scores for all points in inference.}
    \label{Pipeline}
\end{figure*}

\section{Method}
\subsection{Global and Local Registration}
Similar to the Reg3D-AD model \cite{Liu_Xie_Chen_Li_Wang_Liu_Wang_Zheng}, we apply the FPFH \cite{5152473} feature descriptor and the random sample consensus (RANSAC) algorithm to achieve global registration of point clouds. To enhance the stability of point cloud registration, a point-plane ICP \cite{924423} is introduced to locally optimize the outputs from glocal registration. Suppose two point clouds \(X_{s}\) (source point cloud) and \(X_{t}\) (target point cloud) need to be registered, the procedure is as follows:

\begin{enumerate}
\item Apply rotation matrix and translation vector obtained from the global registration to transform \(X_{s}\).
\item Search \(q_{i}\) in \(X_{t}\) such that closest to \(p_{i}\) in \(X_{s}\), where the normal vector of \(q_{i}\) is denoted as \(n_{i}\).
\item Assume optimal rotation Euler angles \(\alpha, \beta, \gamma \rightarrow 0\), we have \(\cos\left(\theta\right) \rightarrow 1, \sin\left(\theta\right) \rightarrow 0, \theta \rightarrow 0\). The rotation matrix R can be approximately expressed as:
\begin{equation}\label{Rotation_equation}
    \begin{matrix}
    R \approx \begin{bmatrix}
    1 & -\gamma & \beta \\
    \gamma & 1 & -\alpha \\
    -\beta & \alpha & 1 \\
    \end{bmatrix}\ \\
    \end{matrix}
\end{equation}

\item Assume the optimal translation vector is \(t = \left[t_{x}, t_{y}, t_{z}\right]\). The loss function is shown as a least squares problem via Moore--Penrose inverse.
\begin{equation}\label{Loss_equation}
   \begin{matrix}
   E\left(R, t\right) = \sum_{i=1}^{n}\left(\left(Rp_{i} + t - q_{i}\right)^{T}n_{i}\right)^{2}\ \\
   \end{matrix}
\end{equation}

\item Apply the computed rotation matrix and translation vector to transform \(X_{s}\), and repeat steps 2-5 until the loss value is below a predefined threshold. Note that the target point cloud \(X_{t}\)  for registration is fixed.
\end{enumerate}

\subsection{Memory Bank Construction}

\textbf{Coordinate Sampling.} %Due to the massive amount of coordinate information in point cloud data, it is necessary to adopt a reasonable method for downsampling the point clouds to accelerate the model's inference speed. In this paper, 
We employ a greedy down-sampling algorithm \cite{7500429} to sample the point clouds. Given the point clouds \(X\), and the point set \(C_{a}\), \(a\) is the number of points in \(C_{a}\). We aim to obtain \(S_{\max}\) uniform points from \(C_{a}\). The specific procedure is as follows:
\begin{enumerate}
\def\labelenumi{\arabic{enumi}.}
\item
  Randomly select \(S_{\text{init}}\) points from \(C_{a}\) to construct the initial point set \(P_{\text{init}} = \left\{ P_{1}, P_{2}, P_{3}, \ldots, P_{S_{\text{init}}} \right\}\).
\item
  Calculate distances between \(C_{a}\) and \(P_{\text{init}}\) to obtain a matrix \(d_{2d}\) with dimensions \(a \times S_{\text{init}}\).
\end{enumerate}
\[d_{2d} = \begin{bmatrix}
d_{11} & \cdots & d_{1S_{\text{init}}} \\
 \vdots & \ddots & \vdots \\
d_{a1} & \cdots & d_{aS_{\text{init}}} \\
\end{bmatrix}\]

\begin{enumerate}
\def\labelenumi{\arabic{enumi}.}
\setcounter{enumi}{2}
\item
  Calculate the mean for each row of the matrix \(d_{2d}\) to obtain \(d_{1d} = \lbrack d_{1_{\text{mean}}},d_{2_{\text{mean}}}\cdots,d_{a_{\text{mean}}}\rbrack\).
\end{enumerate}

\begin{enumerate}
\def\labelenumi{\arabic{enumi}.}
\setcounter{enumi}{3}
\item
  Locate the maximum value in the matrix \(d_{1d}\), and add the corresponding point to \(P_{\text{init}}\). Repeat steps 2-4 until the number of elements in \(P_{\text{init}}\) equals \(S_{\max}\).
\end{enumerate}
\begin{figure}
    \centering
    \includegraphics[width=0.6\linewidth]{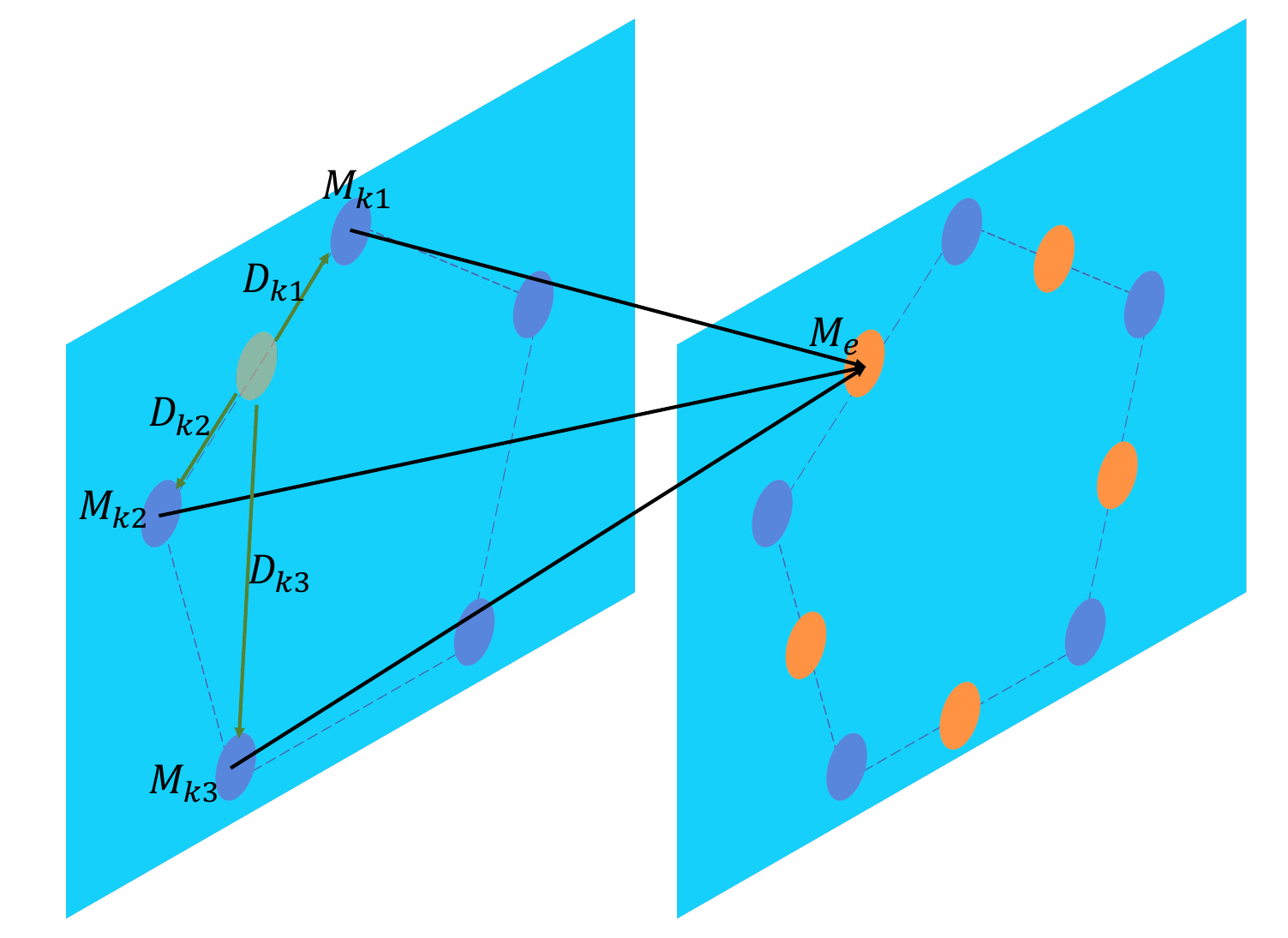}
    \caption{The process of the PointMAE feature interpolation.}
    \label{Interpolation_figure}
\end{figure}

\textbf{Point Feature Interpolation.} A point transformer (PointMAE) \cite{Pang_Wang_Tay_Liu_Tian_Yuan,Zhao_Jiang_Jia_Torr_Koltun_2021} pretrained on the ShapeNet \cite{Chang_Funkhouser_Guibas_Hanrahan_Huang_Li_Savarese_Savva_Song_Su_et} dataset is used as our 3D feature extractor. %For fair comparison, the outputs of layers \{3, 7, 11\} are likewise selected as our 3D feature. The point transformer initially encodes point clouds into point groups, where each group has a center point for position and neighbor numbers for group size.
For each point cloud, we use coordinates from the coordinate sampling stage as the center points of groups. Each central point forms a memory element, which is bound with coordinates and PointMAE features. These elements constitute the memory bank in Fig. \ref{Pipeline}. To reduce the computational complexity of PointMAE features computation in inference, it is necessary to downsample the point cloud coordinates. Therefore, we perform point feature interpolation further to assign a PointMAE feature value to each coordinate in the feature bank. The feature interpolation method is shown in Fig. \ref{Interpolation_figure}.

 Given a point set \(P_{\text{central}}\) with all center point coordinates,  the corresponding PointMAE feature set is \(M_{\text{init}}\). Taking a non-central point \(P_{e}\) as an example, we use the k-nearest neighbors algorithm to obtain three nearest neighbors \(\begin{bmatrix} P_{k1}, P_{k2}, P_{k3}  \end{bmatrix}\) in the point set \(P_{\text{central}}\). The corresponding Euclidean distances and PointMAE feature values are \(\begin{bmatrix} D_{k1}, D_{k2}, D_{k3}  \end{bmatrix}\) and \(\begin{bmatrix} M_{k1}, M_{k2}, M_{k3}  \end{bmatrix}\), respectively. By using eq. \eqref{interpolation_equation}, we obtain the PointMAE feature \(M_{e}\) for \(P_{e}\). This process is repeated until all non-central points obtain their corresponding PointMAE features. The \(M_{e}\) is computed by
\begin{equation}
M_{e} = \frac{D_{k1}D_{k2}M_{k3} + D_{k1}D_{k3}M_{k2} + D_{k2}D_{k3}M_{k1}}{D_{k1}D_{k2} + D_{k1}D_{k3} + D_{k2}D_{k3}} \label{interpolation_equation}
\end{equation}
\begin{table*}[]
\vspace{-2.0em}
\small
\centering
\caption{O-AUROC score for anomaly detection of all categories of Real3D-AD. Our method clearly outperforms other methods in most categories and has a huge advantage in average performance.}
\begin{tabular}{lcc|cc|ccc|cc}
\toprule
\multicolumn{1}{c}{\multirow{2}{*}{\textbf{Category}}} & \multicolumn{2}{c|}{\textbf{BTF \cite{Horwitz_Hoshen_2022}}}                                     & \multicolumn{2}{c|}{\textbf{M3DM \cite{Wang_Peng_Zhang_Yi_Wang_Wang_2023}}}                                         & \multicolumn{3}{c|}{\textbf{PatchCore \cite{Roth_Pemula_Zepeda_Scholkopf_Brox_Gehler_2022}}}                                                                                             & \multicolumn{2}{c}{\textbf{PointCore}}                        \\ \cline{2-10} 
\multicolumn{1}{c}{}                                   & \multicolumn{1}{l}{\textit{Raw}} & \multicolumn{1}{l|}{\textit{FPFH}} & \multicolumn{1}{l}{\textit{PointMAE}} & \multicolumn{1}{l|}{\textit{FPFH}} & \multicolumn{1}{l}{\textit{FPFH+Raw}} & \multicolumn{1}{l}{\textit{PointMAE}} & \multicolumn{1}{l|}{\textit{PointMAE+Raw}} & \textit{FPFH+Raw} & \multicolumn{1}{l}{\textit{PointMAE+Raw}} \\ \hline
Airplane                                               & 0.730                            & 0.520                              & 0.434                                 & \textbf{0.882}                     & 0.848                                 & 0.726                                 & 0.716                                      & 0.792             & 0.660                                     \\
Car                                                    & 0.647                            & 0.560                              & 0.541                                 & 0.590                              & 0.777                                 & 0.498                                 & 0.697                                      & \textbf{0.871}    & 0.866                                     \\
Candybar                                               & 0.539                            & 0.630                              & 0.552                                 & 0.541                              & 0.570                                 & 0.663                                 & 0.685                                      & 0.861             & \textbf{0.976}                            \\
Chicken                                                & 0.789                            & 0.432                              & 0.683                                 & 0.837                              & 0.853                                 & 0.827                                 & \textbf{0.852}                             & 0.842             & 0.841                                     \\
Diamond                                                & 0.707                            & 0.545                              & 0.602                                 & 0.574                              & 0.784                                 & 0.783                                 & 0.900                                      & 0.847             & \textbf{0.963}                            \\
Duck                                                   & 0.691                            & \textbf{0.784}                     & 0.433                                 & 0.546                              & 0.628                                 & 0.489                                 & 0.584                                      & 0.642             & 0.684                                     \\
Fish                                                   & 0.602                            & 0.549                              & 0.540                                 & 0.675                              & 0.837                                 & 0.630                                 & 0.915                                      & 0.915             & \textbf{0.993}                            \\
Gemstone                                               & \textbf{0.686}                   & 0.648                              & 0.644                                 & 0.370                              & 0.359                                 & 0.374                                 & 0.417                                      & 0.477             & 0.535                                     \\
Seahorse                                               & 0.596                            & 0.779                              & 0.495                                 & 0.505                              & 0.767                                 & 0.539                                 & 0.762                                      & 0.954             & \textbf{0.973}                            \\
Shell                                                  & 0.396                            & 0.754                              & 0.694                                 & 0.589                              & 0.663                                 & 0.501                                 & 0.583                                      & 0.853             & \textbf{0.882}                            \\
Starfish                                               & 0.530                            & 0.575                              & 0.551                                 & 0.441                              & 0.471                                 & 0.519                                 & 0.506                                      & 0.617             & \textbf{0.652}                            \\
Toffees                                                & 0.703                            & 0.462                              & 0.450                                 & 0.565                              & 0.626                                 & 0.585                                 & 0.827                                      & 0.728             & \textbf{0.929}                            \\ \hline
Average                                                & 0.635                            & 0.603                              & 0.552                                 & 0.593                              & 0.682                                 & 0.594                                 & 0.704                                      & 0.783           & \textbf{0.829}                            \\ \hline
\end{tabular}
\label{Result_O_AUROC}
\end{table*}

\subsection{Inference Block}

\textbf{Multi-feature Anomaly Score Calculation.} The memory bank is composed of the element sets, that is, \(M_{\text{train}} = \{\left( M_{1_{c}},l M_{1_{p}} \right), \left( M_{2_{c}}, M_{2_{p}} \right), \ldots, \left( M_{n_{c}}, M_{n_{p}} \right)\}\), where \(M_{i_{c}}\) denotes the coordinates of the \(i\)-th point, and \(M_{i_{p}}\) represents the PointMAE feature of the \(i\)-th point. The test feature bank is defined as \(F_{\text{test}} = \{\left( F_{1_{c}}, F_{1_{p}} \right), \left( F_{2_{c}}, F_{2_{p}} \right), \ldots, \left( F_{m_{c}}, F_{m_{p}} \right)\}\), where \(F_{j_{c}}\) is the coordinates of the \(j\)-th point, and \(F_{j_{p}}\) is the PointMAE feature of the \(j\)-th point. For an element \(\left( F_{j_{c}}, F_{j_{p}} \right)\) in \(F_{\text{test}}\), we use its coordinate information \(F_{j_{c}}\) to find three nearest neighbors in \(M_{\text{train}}\), denoted as \(\{\left( M_{i_{c}}, M_{i_{p}} \right), \left( M_{o_{c}}, M_{o_{p}} \right), \left( M_{u_{c}}, M_{u_{p}} \right)\}\). Using Euclidean distance, their coordinate distances \(\{\text{DC}_{1}, \text{DC}_{2}, \text{DC}_{3}\}\) and feature distances \(\{{DP}_{1}, \text{DP}_{2}, \text{DP}_{3}\}\) are obtained. The final coordinate anomaly score is \(S_{c} = \text{mean}\text{DC}_{1}, \text{DC}_{2}, \text{DC}_{3})\) and PointMAE anomaly score is \(S_{p} = \min(\text{DP}_{1}, \text{DP}_{2}, \text{DP}_{3})\).

\begin{figure}[t]
    \subfloat[Traditional normalization method.]{
        \begin{minipage}[t]{0.46\linewidth}
            \centering
            \includegraphics[width=0.95\linewidth]{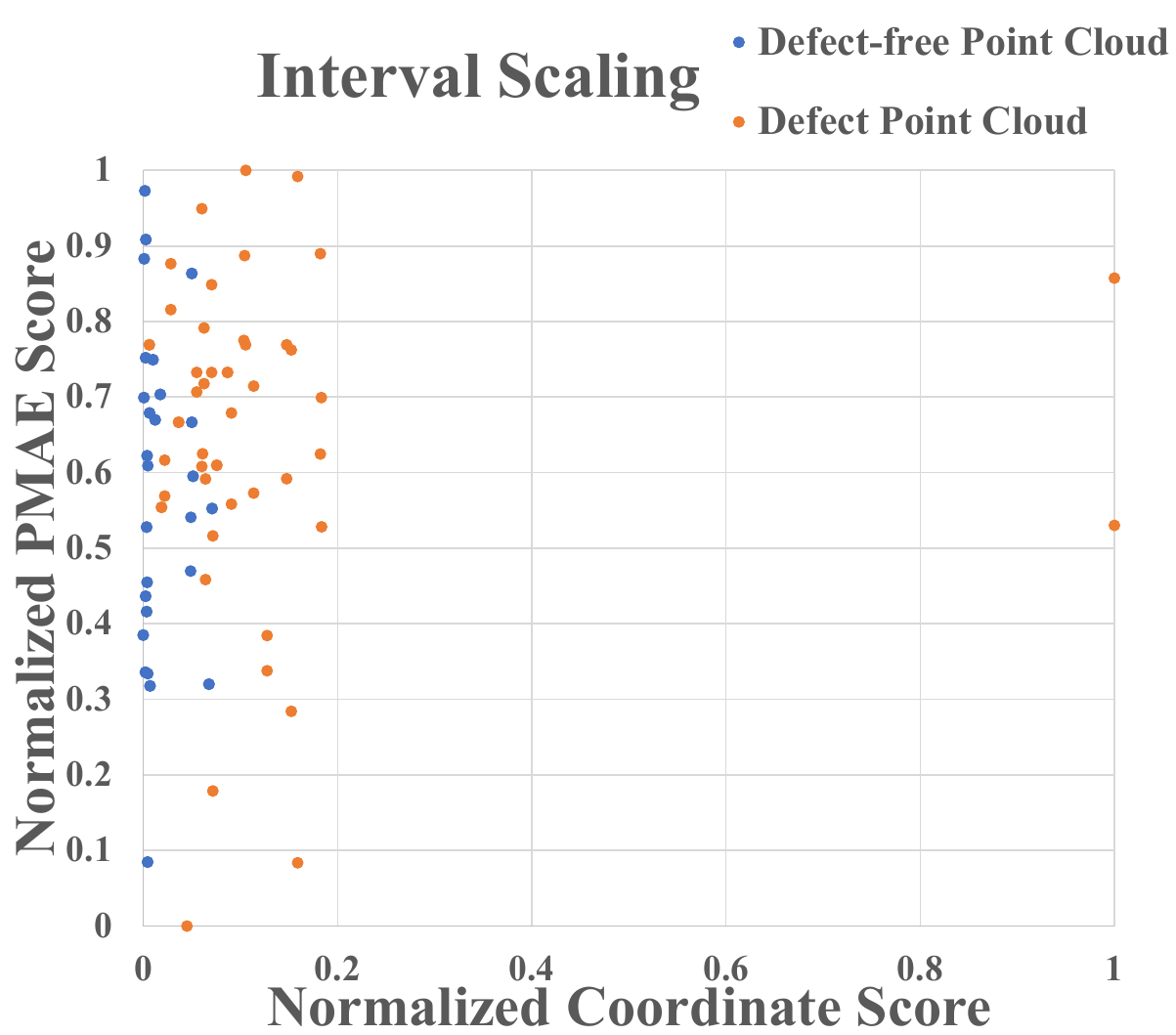}
            \label{fig:hist}
        \end{minipage}
    }
    \hfill
    \subfloat[Ranking-based normalization method.]{
        \begin{minipage}[t]{0.46\linewidth}
            \centering
            \includegraphics[width=0.95\linewidth]{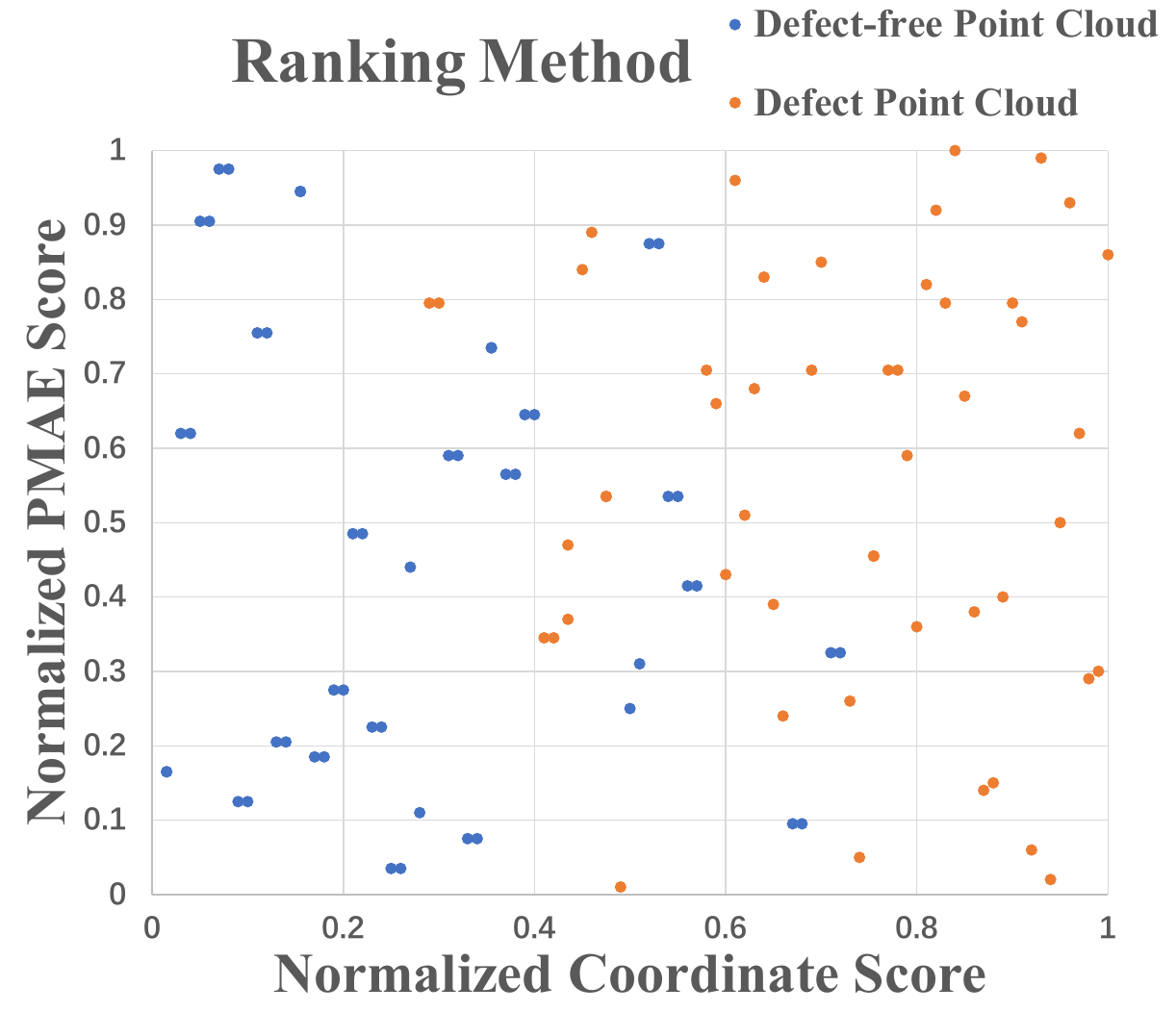}
            \label{fig:tsne}
        \end{minipage}
    }
    \vspace{-5pt}
     \caption{Distribution of coordinate and PointMAE anomaly scores normalized by these two methods.}
      \vspace{-5pt}
     \label{distribution_scores}
\end{figure}

\textbf{Ranking Block}. Due to the differences in scale and distribution between two anomaly scores, both scores are necessary to normalize. Traditional normalization often employs the interval scaling method. For a set of data \(S_{\text{list}}\), the interval scaling process is 
    \(S_{\text{norm}} = \frac{S_{\text{list}} - \min(S_{\text{list}})}{\max(S_{\text{list}})-\min(S_{\text{list}})}\). As depicted in Fig. \ref{distribution_scores}, the interval scaling method will eliminate differences in scale between two anomaly scores but cannot address differences in distribution. When the coordinate anomaly score has two outliers, the final anomaly score will be much smaller than the PointMAE anomaly score. This has a significant impact on ensemble strategies based on arithmetic operations. To address that, we have designed a ranking-based normalization method, where \(Sort\_rank(S_{\text{list}})\) obtains the ranking of each value in \(S_{\text{list}}\), and \(len(S_{\text{list}})\) is the length of \(S_{\text{list}}\). That is:
\begin{equation}\label{ranking_normalization}
    S_{\text{norm}} = \frac{Sort\_rank(S_{\text{list}})}{len(S_{\text{list}})}
\end{equation}

\section{Experiments}

\subsection{Experimental Details}
\textbf{Dataset.} 
Real3D-AD \cite{Liu_Xie_Chen_Li_Wang_Liu_Wang_Zheng} dataset comprises a total of 1,254 samples that are distributed across 12 distinct categories. Each training set for a specific category contains only four samples, similar to the few-shot scenario in 2D anomaly detection. These categories include Airplane, Candybar, Chicken, Diamond, Duck, Fish, Gemstone, Seahorse, Shell, Starfish, and Toffees. All these categories are toys from manufacturing lines. The objects in Real3D-AD dataset have 0.001mm-0.0015mm resolution, 360 degree coverage and perfect prototype. %Noteworthy is its achievement of a point resolution and precision of 0.04mm and 0.011mm, respectively. This is notably higher than other previous point cloud anomaly dataset. Furthermore, the system's strength lies in its implementation of multi-view scanning, effectively mitigating blind spots and enhancing anomaly capture capabilities. Therefore, Real3D-AD is deemed more suitable for achieving high levels of precision in point cloud anomaly detection.

\textbf{Baselines.} We compare with BTF \cite{Horwitz_Hoshen_2022}, M3DM \cite{Wang_Peng_Zhang_Yi_Wang_Wang_2023}, PatchCore \cite{Roth_Pemula_Zepeda_Scholkopf_Brox_Gehler_2022} to evaluate our performance. According to the used point cloud features, they can be grouped into 7 different baselines, namely, BTF (Raw), BTF (FPFH), M3DM (PointMAE), PatchCore (FPFH), PatchCore (FPFH+Raw), PatchCore (PointMAE), PatchCore (PointMAE+RAW), where Raw denotes using coordinate information. PointMAE and FPFH are two different feature descriptors.

\textbf{Evaluation Metrics.} All evaluation metrics are exactly the same as in \cite{Liu_Xie_Chen_Li_Wang_Liu_Wang_Zheng}. We evaluate the object-level anomaly detection performance and the point-level anomaly detection performance via the area under the receiver operator curve (AUROC) and the area under the Precision-Recall curve (AUPR/AP). The higher the AUROC and AUPR, the better anomaly detection performance is. All experiments are conducted on 12th Gen Intel(R) Core(TM) i9-12900K CPU, 64G DDR4 SDRAM and GeForce RTX 3090 platform.

\begin{table*}[]
\vspace{-2.0em}
\small
\caption{Comparing various metrics with the State of the Art (SOTA) model, which is Reg3D-AD (PatchCore+PointMAE+RAW), our model has achieved significant improvements across all metrics while substantially improving the inference speed.}
\centering
% \usepackage{booktabs}
% \usepackage{multirow}
\begin{tabular}{@{}l|ccccc|ccccc@{}}
\toprule
\multicolumn{1}{c}{\multirow{2}{*}{\textbf{Category}}} & \multicolumn{5}{c|}{\textbf{Reg3D-AD}}          & \multicolumn{5}{c}{\textbf{PointCore}}          \\ \cmidrule(l){2-11} 
\multicolumn{1}{c}{}                                   & O-AUROC & P-AUROC & O-AUPR & P-AUPR & Time(s) & O-AUROC & P-AUROC & O-AUPR & P-AUPR & Time(s) \\ \midrule
Airplane                                                & \textbf{0.716}   & \textbf{0.631}   & \textbf{0.703}  & \textbf{0.017}  & 17.759    & 0.66    & 0.608   & 0.667  & 0.016  & \textbf{5.737}     \\
Car                                                     & 0.697   & \textbf{0.718}   & 0.753  & \textbf{0.135}  & 12.938    & \textbf{0.866}   & 0.706   & \textbf{0.862}  & 0.088  & \textbf{4.830}     \\
Candybar                                                & 0.827   & 0.724   & 0.824  & 0.109  & 11.241    & \textbf{0.976}   & \textbf{0.760}   & \textbf{0.973}  & \textbf{0.322}  & \textbf{1.743}     \\
Chicken                                                 & \textbf{0.852}   & 0.676   & \textbf{0.884}  & 0.044  & 18.944    & 0.841   & \textbf{0.780}   & 0.863  & \textbf{0.413}  & \textbf{9.043}     \\
Diamond                                                 & 0.900   & \textbf{0.835}   & 0.884  & 0.191  & 12.611    & \textbf{0.963}   & 0.810   & \textbf{0.957}  & \textbf{0.493}  & \textbf{7.443}     \\
Duck                                                    & 0.584   & 0.503   & 0.588  & 0.01   & 18.014    & \textbf{0.684}   & \textbf{0.712}   & \textbf{0.623}  & \textbf{0.044}  & \textbf{8.987}     \\
Fish                                                    & 0.915   & \textbf{0.826}   & 0.939  & 0.437  & 10.463    & \textbf{0.992}   & 0.782   & \textbf{0.993}  & \textbf{0.510}  & \textbf{1.850}     \\
Gemstone                                                & 0.417   & \textbf{0.545}   & 0.454  & \textbf{0.016}  & 11.108    & \textbf{0.534}   & 0.515   & \textbf{0.548}  & 0.007  & \textbf{3.562}     \\
Seahorse                                                & 0.762   & 0.817   & 0.787  & 0.182  & 9.655     & \textbf{0.973}   & \textbf{0.841}   & \textbf{0.972}  & \textbf{0.637}  & \textbf{1.362}     \\
Shell                                                   & 0.583   & \textbf{0.811}   & 0.646  & 0.065  & 11.091    & \textbf{0.881}   & 0.781   & \textbf{0.774}  & \textbf{0.086}  & \textbf{2.186}     \\
Starfish                                                & 0.506   & 0.617   & 0.491  & 0.039  & 10.145    & \textbf{0.652}   & \textbf{0.736}   & \textbf{0.585}  & \textbf{0.048}  & \textbf{1.652}     \\
Toffees                                                 & 0.685   & \textbf{0.759}   & 0.721  & 0.067  & 12.293    & \textbf{0.929}   & 0.745   & \textbf{0.938}  & \textbf{0.347}  & \textbf{2.990}     \\ \midrule
Average                                                 & 0.704   & 0.705   & 0.723  & 0.109  & 13.022    & \textbf{0.829}   & \textbf{0.731}   & \textbf{0.813}  & \textbf{0.251}  & \textbf{4.282}     \\ \bottomrule
\end{tabular}
\label{compare_sota}
\end{table*}

\subsection{Anomaly Detection on Real3D-AD}

We compare our method with several methods on Real3D-AD, and Table \ref{Result_O_AUROC} shows the anomaly detection results at O-AUROC(object-level AUROC). The FPFH and PointMAE features are combined with Raw feature, respectively. The results indicate that PointMAE-based combination performs better. For the proposed PointCore architecture, the coordinate information of point cloud is indispensable. Table \ref{compare_sota} presents a more comprehensive comparison between our model and the SOTA. PointCore has achieved competitive performance across all metrics, including a 17.75\% improvement in O-AUROC metric. The performance in point-level AUROC (P-AUROC), object-level AUPR (O-AUPR), and point-level AUPR (P-AUPR) further demonstrate the superior performance of our proposal in anomaly detection. %(Memory bank subsampling for all methods is configured at 1\%.)
\renewcommand{\dblfloatpagefraction}{.9}
\begin{table}[]
\small
\center
\caption{Mean inference time per object on Real3D-AD.}
\label{Inference_time}
\begin{tabular}{c|cccc}
\hline
Method     & PointCore           & BTF                 & M3DM                & Reg3D-AD           \\ \hline
O-AUROC     & 0.829         & 0.603         & 0.552         & 0.704        \\
P-AUROC     & 0.731         & 0.571         & 0.637         & 0.705         \\
Time(s)     & 4.282         & 3.882         & 5.061         & 13.022         \\ \hline
\end{tabular}
\end{table}
The comparisons of inference time among BTF, M3DM and PatchCore are implemented, as tabulated in Table \ref{Inference_time}. As can be seen, BTF presented in Table \ref{Inference_time}, albeit fast, show mediocre to poor performance in O-AUROC and P-AUROC. Ours is the fastest excluding BTF.

% \usepackage{multirow}
% \usepackage{multirow}
% \usepackage{multirow}
% \usepackage{multirow}
% \usepackage{booktabs}
% \usepackage{multirow}

% \usepackage{booktabs}
% \usepackage{multirow}
% \usepackage{booktabs}
% \usepackage{multirow}

\renewcommand{\dblfloatpagefraction}{.8}
\begin{table}[]
\vspace{-1.0em}
\caption{The results obtained from ablation experiments on each module. Baseline represents Reg3D-AD, LO(P-P) represents local optimization based on the point-point ICP algorithm, LO represents local optimization based on the point-plane ICP algorithm, and RB represents Ranking Block.}
\center
\includegraphics[width=1\linewidth]{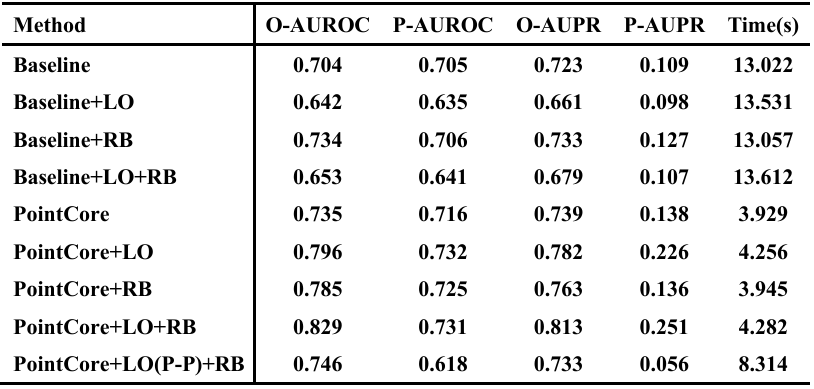}
\label{Ablation}
\vspace{-3.0em}
\end{table}

\subsection{Ablation Studies}
In this section, we ablate our design choices on the Reg3D-AD under 3-view settings in Table. \ref{Ablation}.

\textbf{Effectiveness of Local Optimization (LO).} The Reg3D-AD model employs the FPFH+RANSAC method \cite{5152473} for registration, where the RANSAC algorithm iteratively finds the optimal pose matrix. However, the random points selection of RANSAC in each iteration leads to significant variance in the final registration results. We conducted 20 experiments  on a pair point cloud from the Toffees dataset. The results reveal substantial variations even with identical parameters. The local optimization algorithm is introduced to improve the stability of the registration process. The variance in Euler angles decreased from $1.1058^\circ$ to $(7.7796 \times 10^{- 6})^\circ$,  achieving a score of \(0.642 \pm 0.01\). %While we achieved a more stable model, the performance was slightly worse. However, it is important to note that the decline in metrics is not due to the local optimization method but rather an intrinsic issue in the Reg3D-AD architecture. Through subsequent modifications, we obtained a more stable, faster, and higher-performing model.
Regarding the local optimization method, we conduct registration error tests on point-point ICP and point-plane ICP algorithms under different levels of Gaussian noise. The results show that the point-plane ICP registration method often yields more accurate results under lower noise levels. The anomaly samples in the Reg3D-AD dataset resemble a point cloud with low noise levels. The point-plane ICP is chosen as the local optimization method. 

\textbf{Effectiveness of PointCore Architecture.} Compared to the Reg3D-AD architecture, PointCore architecture can better leverage the coordinate information of point clouds to achieve faster and more accurate point cloud anomaly detection. Regarding speed, we accelerate the inference process by binding the coordinate information and PointMAE feature information of points. This reduces the substantial computational cost introduced by the subsequent PointMAE searching for nearest neighbors. In contrast, the Reg3D-AD architecture stores the coordinate information and PointMAE feature information separately in different memory banks. Each coordinate and PointMAE must find their nearest neighbors in the corresponding memory bank during inference. This poses a significant computational challenge, especially with the 1154 dimensions of PointMAE feature. Regarding accuracy, we enhance the dominance of coordinate information to avoid obvious mismatches in PointMAE features. Specifically, in the Reg3D-AD architecture, the PointMAE features of test point cloud must find the nearest neighbors in the PointMAE memory bank without utilizing any coordinate information. This process leads to erroneous matches between locally similar groups. By strictly limiting the matching range of coordinates, we significantly reduce the probability of mismatches.

\textbf{Effectiveness of Ranking Block (RB).} From Table. \ref{Ablation}, the ranking block significantly improves object-level AUROC and object-level AUPR. However, the improvement in point-level metric is limited. This is because the ranking block is primarily employed to mitigate the substantial impact of outliers on different anomaly score distributions. It plays a significant balancing role for object-level anomaly scores, where the sample size is relatively small. In contrast, point-level anomaly scores inherently have many samples, and the influence of outliers is minimal. %Therefore, the impact of the ranking block is also limited for point-level metrics.

\section{Conclusion}
We propose an unsupervised point cloud anomaly detector, termed PointCore, which is developed on single memory bank with local-global features to store multi-scale information of input point clouds. Extensive experiments on Real3D-AD dataset demonstrate that ours has better recall rate and lower false-positive rate, which is preferable in real applications requiring precise detection of defective samples. Furthermore, the proposed framework is efficient since both the local-global feature memory bank and the multi-feature anomaly score calculation reduce the computational cost.

\bibliographystyle{IEEEtran}
\bibliography{ijcai23}

\end{document}